\documentclass{article} 
\usepackage{iclr2027_conference,times}

\usepackage{amsmath}
\usepackage{libertinust1math}

\usepackage{amsmath,amsfonts,bm}

\def\1{\bm{1}}

\DeclareMathAlphabet{\mathsfit}{\encodingdefault}{\sfdefault}{m}{sl}
\SetMathAlphabet{\mathsfit}{bold}{\encodingdefault}{\sfdefault}{bx}{n}

\def\gL{{\mathcal{L}}}

\usepackage[table]{xcolor}
\usepackage{hyperref}
\usepackage{xurl}
\usepackage{booktabs}
\usepackage{tabularx}
\usepackage{graphicx}
\usepackage{mathtools}
\usepackage{multirow}
\usepackage{subcaption}
\usepackage{tcolorbox}
\usepackage{wrapfig}
\usepackage{needspace}

\newsavebox{\ablationlossbox}

\definecolor{mydarkblue}{rgb}{0,0.08,0.45}
\definecolor{tablehighlight}{HTML}{E1F1FF}
\definecolor{codeborder}{HTML}{53647C}
\hypersetup{
    colorlinks=true,
    linkcolor=mydarkblue,
    filecolor=mydarkblue,
    urlcolor=mydarkblue,
    citecolor=mydarkblue,
}

\newcommand{\aff}[1]{\textsuperscript{#1}}
\newcommand{\tok}[1]{\mathtt{#1}}
\newcommand{\swift}{\textsc{LongSpark}}

\title{Efficient speculative decoding with a fixed-cost parallel drafter}
\author{%
{\small\textbf{Hao-Yuan He\aff{1,2}\thanks{Equal contribution.},\hspace{0.5em}Peng-Fei Liu\aff{4}\footnotemark[1],\hspace{0.5em}Chen Chen\aff{4},\hspace{0.5em}Si Shen\aff{2,3}}} \\
{\small\textbf{Quan He\aff{4},\hspace{0.5em}Runzhou Liu\aff{4},\hspace{0.5em}Xinqi Li\aff{4},\hspace{0.5em}Zhenhua Dong\aff{4},\hspace{0.5em}Ming Li\aff{1,2}}} \\
\aff{1}School of Artificial Intelligence, Nanjing University \\
\aff{2}National Key Laboratory for Novel Software Technology, Nanjing University \\
\aff{3}School of Computer Science, Nanjing University \quad
\aff{4}Huawei Foundation Model Dept \\
}

\iclrfinalcopy 
\begin{document}

\maketitle

\newcommand{\runningtitle}{Efficient Speculative Decoding with a Fixed-Cost Parallel Drafter}
\newbox\runningheadbox
\setbox\runningheadbox=\hbox{\small\bf\runningtitle}
\typeout{RUNNINGHEAD width \the\wd\runningheadbox, textwidth \the\textwidth}
\ifdim\wd\runningheadbox>\textwidth
  \typeout{***********************************************************}
  \typeout{Running head does not fit one line; shorten \string\runningtitle.}
  \typeout{***********************************************************}
  \renewcommand{\runningtitle}{Title Suppressed Due to Excessive Size}
\fi

\newcommand{\headauthors}{}

\renewcommand{\headrulewidth}{0.4pt}

\fancyhf{}
\chead{\small\bf\runningtitle}
\rhead{\headauthors}
\cfoot{\thepage}

\fancypagestyle{preprintfirst}{%
  \fancyhf{}%
  \lhead{\small Preprint}%
  \chead{}
  \rhead{\headauthors}%
  \cfoot{\thepage}%
  \renewcommand{\headrulewidth}{0.4pt}%
}
\thispagestyle{preprintfirst}

\vspace{-0.2in}
\newcommand{\projectpage}{https://long-spark.github.io/}
\newcommand{\coderepo}{https://github.com/Hao-Yuan-He/LongSpark}
\newcommand{\modelrepo}{https://huggingface.co/collections/Hehy/longspark}
{\centering\small
\href{\projectpage}{\includegraphics[height=12pt]{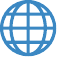}\hspace{0.4em}\textcolor{mydarkblue}{Page}}%
\hspace{1.8em}%
\href{\coderepo}{\includegraphics[height=12pt]{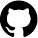}\hspace{0.4em}\textcolor{mydarkblue}{Code}}%
\hspace{1.8em}%
\href{\modelrepo}{\includegraphics[height=12pt]{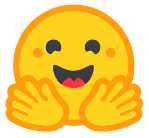}\hspace{0.4em}\textcolor{mydarkblue}{Models}}\par}
\vspace{0.25in}

\begin{abstract}
Speculative decoding accelerates autoregressive inference by verifying multiple draft tokens in a single target forward pass.
However, as the context grows, existing state-of-the-art drafters become increasingly expensive, eroding the very efficiency advantage they are designed to provide.
We argue that this scaling is unnecessary.
A standalone language model must grow with its prefix because it is solely responsible for every token it produces.
A drafter, by contrast, only proposes candidates; the target catches and corrects every error before any token is committed.
The drafter's decoding cost can therefore be made entirely independent of the prefix length.
We introduce \swift{}, a block-diffusion drafter that achieves this by extracting fixed-size, multiscale views from the target's verification pass, thereby eliminating the need for a growing persistent state. Extensive evaluations demonstrate that \swift{} achieves state-of-the-art end-to-end efficiency across multiple model scales and realistic serving conditions. Notably, it delivers the lowest time-per-output-token on long-context tasks while reducing the drafter's context state by several orders of magnitude.
\end{abstract}

\begin{figure}[h]
    \centering
    \includegraphics[width=\linewidth]{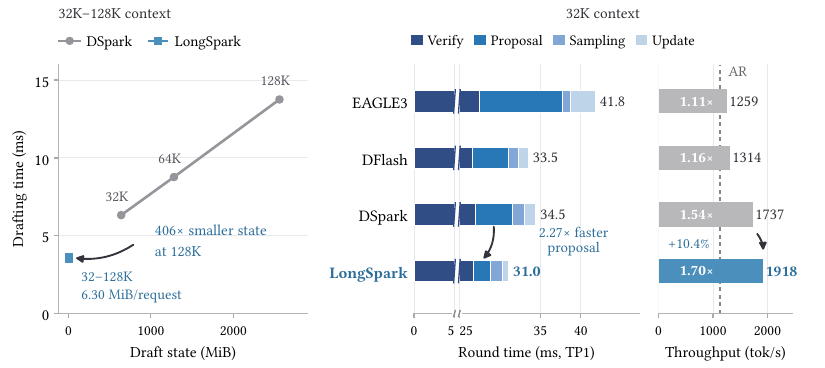}
    \caption{\swift{}: Scalable speculative decoding via fixed-cost drafting.
    Its $\mathcal{O}(1)$ drafting cost keeps drafting time nearly constant as the prefix grows, with $406\times$ smaller drafter context state than full-attention DSpark at 128K tokens (left).
    Drafting time includes proposal and sampling.
    Lower drafting overhead improves end-to-end throughput on \mbox{LongSpec} (right).
    Both panels use Qwen3-8B with concurrency 16.}
    \label{fig:scaling}
\end{figure}
\section{Introduction}

Speculative decoding~\citep{leviathan2023fast} accelerates autoregressive inference by delegating token proposal to a lightweight drafter, which allows a target model to verify multiple candidates in a single forward pass. The core premise of this acceleration is that the drafter must be significantly cheaper than the target for the speedup to be meaningful. However, the cost of producing draft candidates typically scales with the context length, eroding the efficiency advantage exactly when it is most needed.

This scaling bottleneck is systemic across current drafting architectures. Autoregressive drafters maintain KV caches that grow linearly with the prefix; while reusing target representations~\citep{li2024eagle,li2026eagle} can strengthen proposals, it does not eliminate this growing state. Windowed alternatives~\citep{yang2026longspec} bound their own state but still require full-prefix traversals of the target's context. Even advanced block-diffusion drafters, which predict multiple positions jointly~\citep{chen2026dflash,nguyen2026orthrus}, typically inject target features from the entire prefix into every draft layer~\citep{chen2026dflash,cheng2026dspark,huang2026domino,zhang2026dflare}. In each case, the drafter inherits the $\mathcal{O}(t)$ scaling behavior of the target model, transforming the drafter from a lightweight accelerator into a scaling bottleneck in its own right.

But a drafter need not scale this way.
A standalone language model must represent the full prefix faithfully because it bears sole responsibility for every token it produces; any information lost from the prefix becomes an error.
A drafter, by contrast, operates under a different contract: the target verifies every proposal before any token is committed, catching and correcting errors via rejection sampling.
A wrong proposal merely adds a decoding round, never an incorrect output.
The drafter can therefore work from a compressed, lossy view of the prefix.

This observation shifts the fundamental design objective for drafting: the true measure of efficiency is the number of accepted tokens produced relative to the drafting overhead. When a drafter maintains sufficient context to generate useful proposals at a constant cost, its per-round overhead decouples from the prefix length---a principle we term \emph{fixed-cost drafting}.

We achieve this with \swift{}, a block-diffusion drafter. It extracts all context from the target's verification pass as fixed-size, multiscale views: a single-position anchor at the decoding boundary, a short window of recent token-level detail, and a global context summary that compresses the entire prefix into a fixed number of entries. By integrating these views into a parallel block-prediction framework, \swift{} ensures that target verification is the only full-prefix traversal in each round. Every other operation---from context extraction to iterative refinement---remains strictly $O(1)$ relative to the confirmed prefix.

We evaluate \swift{} across reasoning, coding, and dialogue benchmarks at multiple model scales, confirming that it achieves the highest end-to-end throughput, delivering $1.88\times$ to $2.13\times$ speedups that grow with target size. Indeed, \swift{} outperforms state-of-the-art drafters in throughput even when they achieve longer accepted lengths, validating that strictly bounded overhead is more critical for overall latency than marginal gains in proposal quality. Across three long-context benchmarks up to $128\text{K}$ tokens and all three target scales, \swift{} delivers the lowest mean time per output token while reducing drafter state by several orders of magnitude.

The contributions of this work are:
\begingroup
\setlength{\leftmargini}{15pt}
\setlength{\labelsep}{5pt}
\setlength{\labelwidth}{10pt}
\begin{itemize}
    \item We identify an asymmetry between drafter and target: because verification guarantees output correctness, the drafter's decoding cost need not scale with the prefix length. We formalize this as \emph{fixed-cost drafting}.
    \item We introduce \swift{}, a block-diffusion drafter that extracts all of its context as fixed-size, multiscale views of the target's verification pass, making every additional per-round operation $O(1)$ with respect to the prefix length.
    \item We empirically validate fixed-cost drafting across diverse benchmarks and model scales, demonstrating that \swift{} improves end-to-end throughput while reducing drafter memory overhead by several orders of magnitude.
\end{itemize}
\endgroup

\section{\swift{}: Fixed-Cost Speculative Drafting}\label{sec:method}

Fixed-cost drafting requires two ingredients: a context interface that extracts bounded-size views from an unbounded prefix, and a proposal model that produces competitive drafts from these views alone. We formalize both below, after establishing the necessary background (\S~\ref{sec:background}). The context interface (\S~\ref{sec:context}) and proposal model (\S~\ref{sec:proposal}) are followed by training, inference, and complexity analysis (\S~\ref{sec:training}).

\begin{figure}[t]
    \centering
    \includegraphics[width=.96\linewidth]{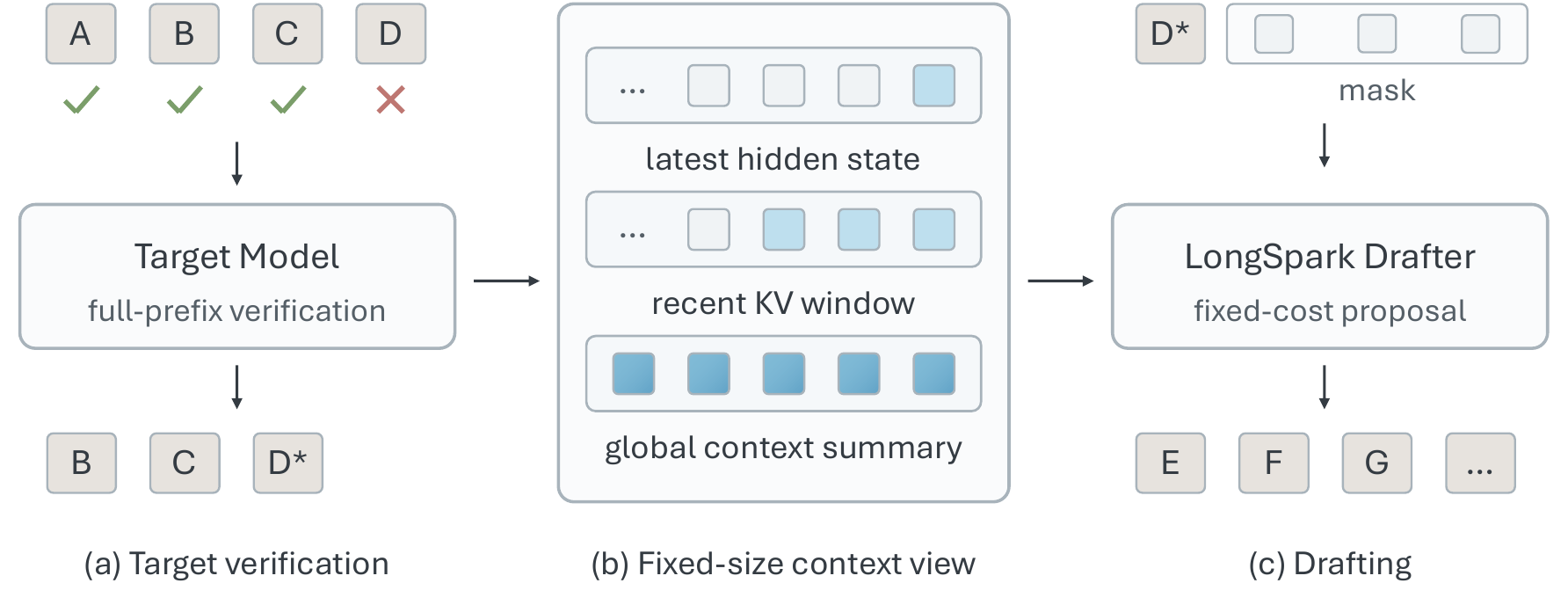}
    \caption{One decoding round of \swift{}. (a) The target verifies the proposed tokens. (b) Three fixed-size context views are extracted from the target's state: a boundary state, a recent KV window, and a global context summary. (c) The drafter uses these views to propose the next token block at a cost independent of prefix length.}
    \label{fig:pipeline}
\end{figure}
\subsection{Preliminaries}\label{sec:background}

\paragraph{Speculative decoding.}
Speculative decoding~\citep{leviathan2023fast} pairs a target model with a lightweight drafter. At the start of a round, the target has cached $x_{1:t}$ but has yet to process its latest committed token $\tok{a}=x_{t+1}$. Given the committed prefix $s=(x_{1:t},\tok{a})$, the drafter proposes $B$ candidates $y_{1:B}$ by sampling from distributions $q_i(\cdot)=q(\cdot\mid s,y_{<i})$. The target verifies these by processing $[\tok{a},y_{1:B}]$ in a single forward pass to compute $p_i(\cdot)=p(\cdot\mid s,y_{<i})$. Candidates are tested in order, accepting $y_i$ with probability
\begin{equation}
\alpha_i=\min\!\left(1,\frac{p_i(y_i)}{q_i(y_i)}\right).
\label{eq:acceptance}
\end{equation}
At the first rejected position $i$, we discard $y_{i:B}$ and sample a correction token from
\begin{equation}
p_i^{\mathrm{corr}}(v)=\frac{[p_i(v)-q_i(v)]_+}{\sum_{u}[p_i(u)-q_i(u)]_+},
\qquad [z]_+=\max(z,0).
\label{eq:residual}
\end{equation}
Each round adds one token beyond the accepted candidates: a correction token upon rejection, or a sample from $p(\cdot\mid s,y_{1:B})$ when all $B$ candidates are accepted. Accepting $r$ candidates therefore advances the prefix by $r+1$ tokens while preserving the target distribution. With mean drafting and verification latencies $T_{\mathrm{draft}}$ and $T_{\mathrm{verify}}$ and mean advancement $\tau=\mathbb{E}[r+1]$, the average decoding time per token is modeled as
\begin{equation}
L=\frac{T_{\mathrm{draft}}+T_{\mathrm{verify}}}{\tau}.
\label{eq:decoding-latency}
\end{equation}

\paragraph{Block-diffusion drafting.}
Single-step block-diffusion drafters such as DFlash~\citep{chen2026dflash} reduce $T_{\mathrm{draft}}$ by predicting a block of proposal distributions in one denoising pass. Let $\mathcal C(s)$ denote context features extracted from the target and $\bm u_{1:B}$ denote input representations for the proposal slots, such as mask embeddings or available committed-token embeddings. A bidirectional draft network $F_\theta$ produces base logits jointly:
\begin{equation}
\bm z_{1:B}^{(0)}=F_\theta(\bm u_{1:B};\mathcal C(s)),
\qquad
q_i^{(0)}=\operatorname{softmax}(\bm z_i^{(0)}).
\label{eq:block-diffusion}
\end{equation}
Sampling each position from these distributions yields the factorized proposal $q^{(0)}(y_{1:B}\mid s)=\prod_{i=1}^{B}q_i^{(0)}(y_i\mid s)$. The backbone mixes information across positions through bidirectional attention, while tokens are sampled independently from its outputs. Semi-autoregressive variants~\citep{cheng2026dspark} add lightweight token dependencies after this parallel pass. The cost of this pass still grows with prefix length when attention spans the full prefix. The remaining question is whether a fixed-size context view $\mathcal C(s)$ can retain competitive proposal quality.

\subsection{Fixed-Cost Context Interface}\label{sec:context}

We instantiate $\mathcal C(s)$ with three fixed-size views that capture the target's cached state at complementary temporal scales. A \emph{boundary state} anchors the current generation point, fusing target hidden representations $\{\bm h_t^\ell\}_{\ell\in\gL}$ at the decoding boundary into a single conditioning vector $\bm c_t$ via learned projection. A \emph{recent KV window} of $W$ entries per selected layer preserves token-level detail in the local neighborhood. A \emph{global context summary} compresses the entire prefix into $R$ entries per layer and head. All three are extracted from a subset $\gL$ of evenly spaced target layers (Appendix~\ref{app:arch}) and involve only constant-size reads and transformations.

\begin{figure}[t]
    \centering
    \includegraphics[width=.9\linewidth]{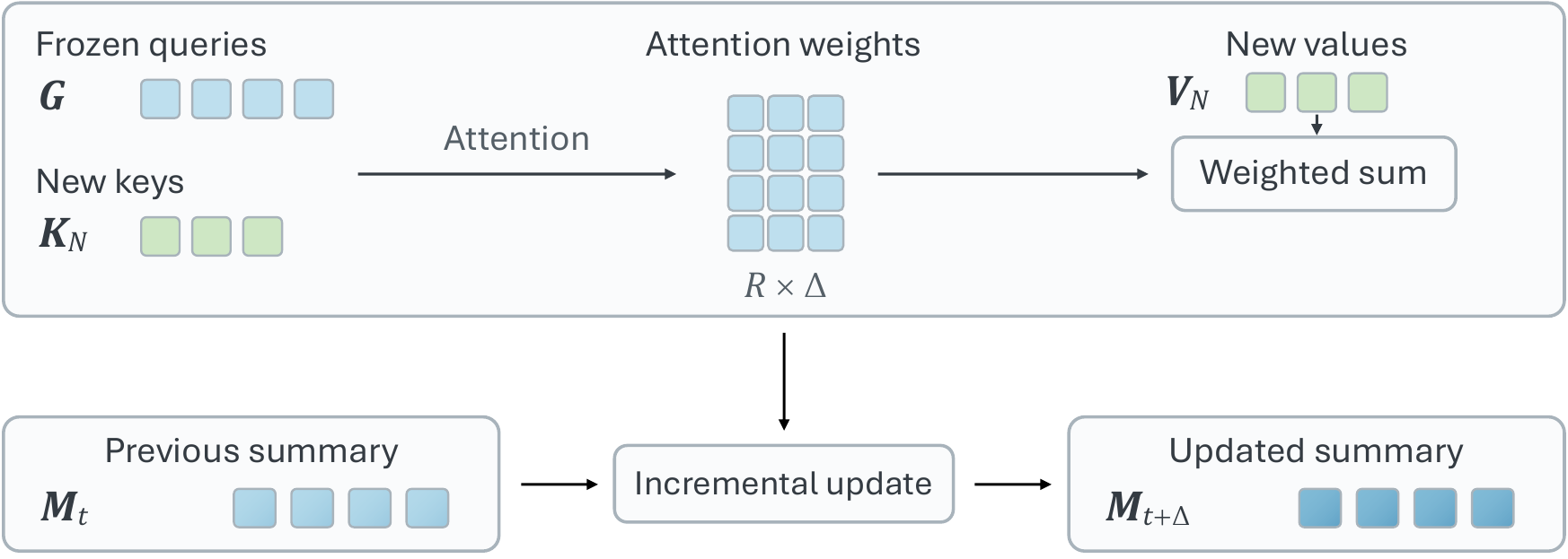}
    \caption{Incremental global context summary during inference. The summary is initialized over the full prefix during target prefill, then updated using only newly retained KV rows $N$ ($\Delta=|N|\leq B+1$).}
    \label{fig:global-summary}
\end{figure}

\paragraph{Global context summary.}
The summary must compress the entire prefix into a fixed number of entries while remaining incrementally updatable as new tokens are committed. To achieve this, each selected layer and attention head maintains a bank of $R$ learned summary queries $\bm{G} \in \mathbb{R}^{R \times d}$, which are independent of the prefix and held fixed during inference (Appendix~\ref{app:training}). For a head of dimension $d$ and corresponding target keys and values $\bm{K}_{1:t}$ and $\bm{V}_{1:t}$, the summary values are computed as
\begin{equation}
\bm{M}_t=\operatorname{Attn}(\bm{G},\bm{K}_{1:t},\bm{V}_{1:t}).
\label{eq:global-summary}
\end{equation}
Because $\bm{G}$ remains fixed during inference, historical attention contributions remain valid as the context grows, and only the newly retained target KV rows need to be incorporated. After verification, the summary absorbs at most $B+1$ new rows (Figure~\ref{fig:global-summary}), keeping update cost and stored state independent of $t$. The drafter reads $(\bm{G},\bm{M}_t)$ as summary keys and values, respectively: the same queries that produced the summary now serve as retrieval keys for the draft layers. Appendix~\ref{app:summary-update} gives the incremental update algorithm.

\subsection{Proposal Model}\label{sec:proposal}

We instantiate the parallel backbone in \eqref{eq:block-diffusion} with the fixed-cost interface from \S~\ref{sec:context}, then introduce token-level dependencies through a lightweight sequential correction. The result is a $D$-layer Transformer that processes all $B$ draft positions in parallel, followed by a per-position Markov correction that conditions each token on its predecessor.

The backbone operates on $B$ input slots, where the first encodes the committed token $\tok{a}$ and the remaining $B-1$ employ mask-token embeddings. Their embeddings are combined with within-block positional encodings and the boundary state $\bm{c}_t$ to initialize the draft representations. The output of slot $i$ supplies the base logits for proposal $y_i$, for $i=1,\ldots,B$ (Appendix~\ref{app:drafter-forward}).

These representations are processed by $D=|\gL|$ Transformer layers, where each draft layer $j$ is paired with a corresponding selected target layer $\ell(j)$, from which it reads the recent window and global summary. At each layer, attention is computed over a concatenated context:
\begin{equation}
\begin{aligned}
\bm{K} &= [\bm{K}^{\mathrm b};\;\bm{K}^{\mathrm w};\;\bm{G}],\\
\bm{V} &= [\bm{V}^{\mathrm b};\;\bm{V}^{\mathrm w};\;\bm{M}_t],
\end{aligned}
\label{eq:draft-context}
\end{equation}
where superscripts $\mathrm b$ and $\mathrm w$ denote the proposal block and recent window, respectively; layer and head indices are omitted for readability. Consequently, all block positions attend to one another through a fixed-size context of at most $B+W+R$ entries per head, ensuring the computation remains independent of the prefix length.

Following the $D$ draft layers, the frozen Target normalization and vocabulary head generate base logits $\bm{z}_i^{(0)}$ for all $B$ positions in parallel. These are then refined by a low-rank sequential correction~\citep{cheng2026dspark} that conditions each position on the token sampled at its predecessor. With $y_0=\tok{a}$, a correction embedding $\bm{E}_{\mathrm c}$, and an output mapping $\bm{W}_{\mathrm c}$, the final logits are given by:
\begin{equation}
\bm{z}_i
=
\bm{z}_i^{(0)}+\bm{W}_{\mathrm c}\bm{E}_{\mathrm c}(y_{i-1}),
\qquad i=1,\ldots,B.
\label{eq:markov}
\end{equation}
The resulting proposal factorizes as
\begin{equation}
q(y_{1:B}\mid s)=\prod_{i=1}^{B}q_i(y_i\mid s,y_{i-1}),
\qquad q_i=\operatorname{softmax}(\bm z_i),
\label{eq:proposal-factorization}
\end{equation}
with $y_0=\tok{a}$. These corrected distributions are used for both sampling and verification in \eqref{eq:acceptance}--\eqref{eq:residual}. Only the correction and sampling occur sequentially; the Transformer backbone and base vocabulary projection remain fully parallel.

\subsection{Training and Inference}\label{sec:training}\label{sec:verify}

\paragraph{Training.}
All Target parameters, including the token embedding, normalization, and vocabulary head, remain frozen. We jointly train the draft network, boundary fusion and input mappings, summary queries, and the DSpark-style sequential correction~\citep{cheng2026dspark}. The full-vocabulary distributions $p_i$ and $q_i$ are evaluated under teacher forcing: the target sees the ground-truth prefix $x_{1:t+i}$, and the correction in \eqref{eq:markov} uses the predecessor $x_{t+i}$. For ground-truth token $x_{t+1+i}$, the weighted loss at position $i$ is
\begin{equation}
\mathcal J_i
=
w_i\Bigl[\lambda\bigl[-\log q_i(x_{t+1+i})\bigr]
+(1-\lambda)\lVert q_i-p_i\rVert_1\Bigr],
\label{eq:training-objective}
\end{equation}
where $\lambda=0.1$ and $w_i=\exp(-(i-1)/4)$. We sum these losses over valid training positions and normalize by the sum of their weights. At inference, the correction instead uses the previously sampled token.

\paragraph{Inference.}
Verification follows \S~\ref{sec:background}, using the corrected proposal distributions from \eqref{eq:proposal-factorization}. When $r$ candidates are accepted, the prefix advances by $r+1$ positions, and the committed token $\tok{a}$ along with $y_{1:r}$ enter the Target KV cache. This update naturally drives the incremental maintenance of the interface: their KV rows update the global context summary and advance the recent window, while the hidden states at the new boundary yield the updated $\bm{c}_t$. The process concludes by setting the correction or bonus token as the next committed token $\tok{a}$ and discarding all transient draft KV, leaving the updated interface ready for the next round.

\paragraph{Fixed-cost complexity.}
For fixed model dimensions, each draft layer attends to $B$ block positions, $W$ recent tokens, and $R$ summary entries, costing $O(B^2+BW+BR)$ per round. Full-prefix draft attention instead costs $O(B^2+Bt)$. Interface maintenance reads a constant-size boundary and window and incorporates at most $B+1$ newly retained KV rows into the summary. With $H$ attention heads of dimension $d$, the summary occupies $O(|\gL|HRd)$ persistent state; the recent window is read directly from the target cache, and block KV is transient. Both the drafter's additional state and its per-round computation are therefore independent of $t$. Full-prefix processing is confined to target verification, with the global summary initialized once during prefill.

\section{Experiments}\label{sec:experiments}
We organize the evaluation around three questions.
Does a fixed-cost context interface outperform drafters that read the entire prefix across model scales, tasks, and serving loads (\S~\ref{sec:overall})?
Does its advantage persist as contexts lengthen, with drafting cost independent of the prefix length (\S~\ref{sec:scaling})?
And how much does each context component contribute (\S~\ref{sec:module-analysis})?

\subsection{Experimental Setup}\label{sec:setup}

\paragraph{Models and baselines.}
To demonstrate the generalizability and scalability of \swift{}, we evaluate it across three Target scales: Qwen3-4B, 8B, and 14B. We compare our approach against a suite of representative baselines, including the autoregressive EAGLE-3~\citep{li2026eagle} and state-of-the-art parallel drafters DFlash~\citep{chen2026dflash} and DSpark~\citep{cheng2026dspark}, alongside vanilla autoregressive decoding.

\paragraph{Training data.}
We train \swift{} on Open-PerfectBlend~\citep{labonne2024perfectblend}, a high-quality instruction-tuning dataset of approximately 1.42 million samples, for 10 epochs. We use only the prompts and regenerate all responses with the corresponding Target model, ensuring the drafter learns the exact distribution it is intended to accelerate.  Training details are provided in Appendix~\ref{app:training}.

\paragraph{Evaluation benchmarks.}
We subject \swift{} to a comprehensive set of stress tests across eight benchmarks spanning three primary domains: mathematical reasoning, i.e., GSM8K~\citep{cobbe2021gsm8k}, MATH-500~\citep{lightman2023verify}, and AIME25~\citep{maa2025aime}, code generation, i.e., MBPP~\citep{austin2021mbpp}, HumanEval~\citep{chen2021humaneval}, and LiveCodeBench (LCB)~\citep{jain2024livecodebench}, and open-ended dialogue, i.e., MT-Bench~\citep{zheng2023mtbench} and Alpaca~\citep{taori2023alpaca}. The main long-context evaluation covers continuation tasks from LongSpec's training corpus~\citep{yang2026longspec} at 32K, our curated CodeSpan at 64K, and LongSWE-Bench (LSWE)~\citep{rando2025longcodebench} at 128K. Details of the long-context datasets are provided in Appendix~\ref{app:codespan}.

\paragraph{Evaluation settings and metrics.}
We use a target--drafter (TD) disaggregated framework: speculative methods add one dedicated drafter GPU to the Target GPUs used by autoregressive decoding. Main-text results use $T=1$, thinking mode disabled, and seven draft tokens per verification round for speculative methods. We report throughput speedup over autoregressive decoding and average accepted length $\tau$, with output throughput (TPS) and time per output token (TPOT) for long-context workloads. Detailed metric definitions are provided in Appendix~\ref{app:measurement}.

\begin{table}[t]
\centering
\captionsetup{skip=6pt}
\caption{Speculative decoding across model scales and tasks. Entries report accepted length $\tau$, with throughput speedup over autoregressive decoding in parentheses. Avg.\ denotes the mean across eight benchmarks; bold marks the highest speedup.}
\label{tab:main-results}
\scriptsize
\setlength{\tabcolsep}{0.75pt}
\renewcommand{\arraystretch}{1.20}
\begin{tabularx}{\linewidth}{@{}c>{\hspace{3pt}}l*{9}{>{\centering\arraybackslash}X}@{}}
\toprule
\multirow{2}{*}{Target} & \multirow{2}{*}{Drafter} & \multicolumn{3}{c}{\textbf{Math}} & \multicolumn{3}{c}{\textbf{Code}} & \multicolumn{2}{c}{\textbf{Chat}} & \textbf{Overall} \\
\cmidrule(lr){3-5} \cmidrule(lr){6-8} \cmidrule(lr){9-10} \cmidrule(lr){11-11}
 & & GSM8K & MATH-500 & AIME25 & MBPP & HumanEval & LCB & MT-Bench & Alpaca & Avg. \\
\midrule
\multirow{4}{*}{Q3-4B} & EAGLE-3 & 5.15 (1.36) & 4.60 (1.35) & 3.84 (1.24) & 3.69 (1.03) & 4.15 (1.17) & 3.77 (1.14) & 2.42 (0.72) & 2.26 (0.69) & 3.74 (1.09) \\
 & DFlash & 5.37 (1.81) & 4.89 (1.89) & 4.01 (1.72) & 4.40 (1.55) & 4.74 (1.72) & 4.23 (1.61) & 3.05 (1.18) & 2.94 (1.14) & 4.20 (1.58) \\
 & DSpark & 6.10 (1.99) & 5.72 (2.14) & 4.92 (2.05) & 5.14 (1.73) & 5.44 (1.91) & 4.92 (1.79) & 3.65 (1.36) & 3.55 (1.32) & 4.93 (1.79) \\
\rowcolor{tablehighlight}
\cellcolor{white} & \swift{} & 5.90 (\textbf{2.08}) & 5.52 (\textbf{2.28}) & 4.67 (\textbf{2.17}) & 5.01 (\textbf{1.84}) & 5.12 (\textbf{1.97}) & 4.63 (\textbf{1.86}) & 3.52 (\textbf{1.46}) & 3.46 (\textbf{1.42}) & 4.73 (\textbf{1.88}) \\
\midrule
\multirow{4}{*}{Q3-8B} & EAGLE-3 & 5.26 (1.49) & 4.75 (1.50) & 4.02 (1.37) & 3.92 (1.16) & 4.32 (1.30) & 4.17 (1.28) & 2.67 (0.85) & 2.54 (0.82) & 3.96 (1.22) \\
 & DFlash & 5.36 (1.96) & 4.91 (2.02) & 4.07 (1.83) & 4.36 (1.66) & 4.67 (1.82) & 4.44 (1.68) & 3.09 (1.28) & 3.00 (1.24) & 4.24 (1.69) \\
 & DSpark & 6.17 (2.17) & 5.80 (2.32) & 4.99 (2.19) & 5.18 (1.89) & 5.48 (2.06) & 5.09 (1.84) & 3.69 (1.47) & 3.59 (1.42) & 5.00 (1.92) \\
\rowcolor{tablehighlight}
\cellcolor{white} & \swift{} & 5.93 (\textbf{2.24}) & 5.57 (\textbf{2.42}) & 4.72 (\textbf{2.27}) & 5.03 (\textbf{1.98}) & 5.16 (\textbf{2.11}) & 4.83 (\textbf{1.88}) & 3.53 (\textbf{1.54}) & 3.46 (\textbf{1.49}) & 4.78 (\textbf{1.99}) \\
\midrule
\multirow{4}{*}{Q3-14B} & EAGLE-3 & 5.22 (1.69) & 4.62 (1.63) & 3.82 (1.44) & 3.81 (1.29) & 4.11 (1.41) & 4.01 (1.35) & 2.61 (0.93) & 2.49 (0.90) & 3.84 (1.33) \\
 & DFlash & 5.37 (2.16) & 4.86 (2.17) & 4.01 (1.91) & 4.45 (1.85) & 4.57 (1.94) & 4.35 (1.72) & 3.09 (1.38) & 2.96 (1.33) & 4.21 (1.81) \\
 & DSpark & 6.19 (2.43) & 5.77 (2.53) & 4.93 (2.32) & 5.25 (2.12) & 5.41 (2.24) & 4.97 (1.88) & 3.67 (1.61) & 3.56 (1.56) & 4.97 (2.09) \\
\rowcolor{tablehighlight}
\cellcolor{white} & \swift{} & 6.03 (\textbf{2.47}) & 5.63 (\textbf{2.61}) & 4.76 (\textbf{2.37}) & 5.14 (\textbf{2.18}) & 5.15 (\textbf{2.25}) & 4.78 (\textbf{1.90}) & 3.57 (\textbf{1.65}) & 3.48 (\textbf{1.60}) & 4.82 (\textbf{2.13}) \\
\bottomrule
\end{tabularx}
\end{table}

\subsection{Overall Performance}\label{sec:overall}

We first evaluate end-to-end efficiency across model scales and task domains, using target TP1, concurrency 32, and up to 2,048 generated tokens per request. As shown in Table~\ref{tab:main-results}, \swift{} achieves the highest average throughput at every scale, delivering $1.88\times$, $1.99\times$, and $2.13\times$ speedups from 4B to 14B---with the speedup growing as the target scales. Crucially, it keeps this lead even when DSpark attains longer accepted lengths. This separation exposes the central tension of speculative decoding: end-to-end efficiency depends not on proposal quality alone, but on the balance between accepted tokens and the overhead of producing them. The same ranking holds under greedy decoding, confirming that the advantage of a fixed-cost drafter is structural rather than sampling-dependent (Appendix~\ref{app:accepted-length}).

\begin{figure}[tbp]
\centering
\includegraphics[width=0.98\linewidth]{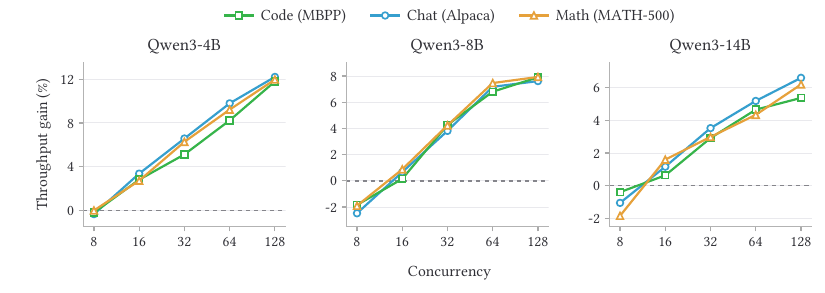}
\caption{Higher concurrency increases serving load. As this load rises, \swift{}'s throughput advantage over DSpark grows.}
\label{fig:concurrency}
\end{figure}

\paragraph{High-concurrency performance.}
To examine how the advantage scales with serving load, we vary concurrency from 8 to 128 on Alpaca, MBPP, and MATH-500. As Figure~\ref{fig:concurrency} shows, all three target scales exhibit the same pattern: \swift{} is marginally behind DSpark at low concurrency, then overtakes it and pulls away steadily as load rises, leading by a clear advantage at concurrency 128. The gap widens on every benchmark despite shorter accepted lengths, revealing that the value of low-overhead drafting compounds under high serving load. Appendix~\ref{app:concurrency} provides additional concurrency results.

\subsection{Scaling with Context Length}\label{sec:scaling}

\paragraph{Long-context performance.}
We evaluate contexts from 32K to 128K using Qwen3-8B with target TP4, concurrency 16, and up to 8,192 generated tokens per request. Long-context evaluation uses fixed NTK scaling~\citep{bloc972023ntk} with $\alpha=4$. As shown in Table~\ref{tab:long-context}, \swift{} achieves the highest mean throughput and lowest mean TPOT across all three workloads. On CodeSpan at 64K, its mean TPS exceeds that of DSpark, the strongest competing method, by 29.1\%, while reducing mean TPOT by 23.1\%. Results for Qwen3-4B and 14B are provided in Appendix~\ref{app:longcontext-scales}.

\begin{table}[t]
\centering
\captionsetup{skip=6pt}
\caption{Long-context decoding on Qwen3-8B. We report TPS (tokens/s) and TPOT (ms/token), averaged over ten seeds. Bold marks the best mean.}
\label{tab:long-context}
\small
\renewcommand{\arraystretch}{1.18}
\setlength{\tabcolsep}{4pt}
\begin{tabularx}{\linewidth}{l*{6}{>{\centering\arraybackslash}X}}
\toprule
\multirow{2}{*}{Method} & \multicolumn{2}{c}{LongSpec 32K} & \multicolumn{2}{c}{CodeSpan 64K} & \multicolumn{2}{c}{LSWE 128K} \\
\cmidrule(lr){2-3} \cmidrule(lr){4-5} \cmidrule(lr){6-7}
 & TPS $\uparrow$ & TPOT $\downarrow$ & TPS $\uparrow$ & TPOT $\downarrow$ & TPS $\uparrow$ & TPOT $\downarrow$ \\
\midrule
Vanilla & 1128.9 & 14.0 & 884.4 & 17.8 & 160.5 & 97.2 \\
EAGLE-3 & 1258.5 & 12.5 & 1071.8 & 14.7 & 158.7 & 97.4 \\
DFlash & 1313.7 & 12.0 & 919.1 & 17.2 & 163.6 & 94.6 \\
DSpark & 1736.6 & 9.0 & 1217.3 & 13.0 & 190.9 & 81.4 \\
\rowcolor{tablehighlight}
\swift{} & \textbf{1917.6} & \textbf{8.2} & \textbf{1571.8} & \textbf{10.0} & \textbf{212.7} & \textbf{74.3} \\
\bottomrule
\end{tabularx}
\end{table}

\paragraph{Drafting cost and memory.}
The throughput gains of \swift{} stem from its lean drafting overhead. Figure~\ref{fig:scaling} (right) decomposes a decoding round on LongSpec with Qwen3-8B into proposal and verification. Target verification dominates the round, and \swift{} nearly halves the proposal component relative to DSpark. Because the round is bounded by verification, this reduction lowers total round latency only modestly; its value lies in repetition, as the saving accumulates over the thousands of rounds needed to generate a long response.

The context-length sweep in Figure~\ref{fig:scaling} (left) confirms that the fixed-size interface decouples the drafter's decoding cost from the prefix length, allowing the advantage to grow with context rather than fade. As the context grows from 32K to 128K, full-attention DSpark's drafting time rises from 6.3 to 13.8\,ms and its effective context state grows from 640 to 2560\,MiB, since both quantities are tied to the number of retained tokens. \swift{} holds both nearly constant: its drafting time stays near 3.5\,ms and its context state remains fixed at 6.3\,MiB across the same range, amounting to a $406\times$ reduction in drafter state at 128K.

These measurements isolate where the cost of drafting actually resides. Since target verification remains the only full-prefix operation in each round, the drafter's contribution can be made both small and constant. This is the structural distinction that separates \swift{} from prefix-scaling drafters: for them, a longer context translates directly into more work per proposal, whereas here the proposal cost is a fixed tax that never grows.

\subsection{Ablation study}\label{sec:module-analysis}

\begin{figure}[t]
\centering
\sbox{\ablationlossbox}{\includegraphics[width=0.46\linewidth]{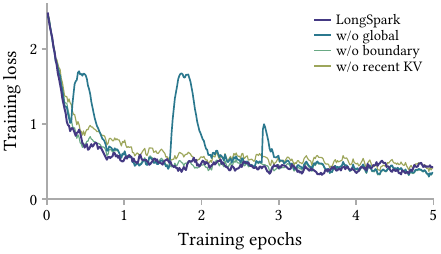}}
\captionsetup[subfigure]{position=top,skip=6pt}
\begin{subfigure}[t]{0.46\linewidth}
\centering
\subcaption{Accepted length $\tau$.}
\begin{minipage}[c][\dimexpr\ht\ablationlossbox+\dp\ablationlossbox\relax][c]{\linewidth}
\centering
\footnotesize
\renewcommand{\arraystretch}{1.35}
\setlength{\tabcolsep}{2.4pt}
\begin{tabularx}{\linewidth}{l*{4}{>{\centering\arraybackslash}X}}
\toprule
Variant & Math & Code & Chat & Avg. \\
\midrule
\swift{} & \textbf{5.33} & \textbf{4.91} & \textbf{3.40} & \textbf{4.69} \\
w/o global summary & 4.99 & 4.50 & 3.17 & 4.35 \\
w/o boundary state & 4.90 & 4.42 & 3.15 & 4.28 \\
w/o recent KV window & 4.17 & 3.82 & 2.86 & 3.71 \\
\bottomrule
\end{tabularx}
\end{minipage}
\end{subfigure}\hfill
\begin{subfigure}[t]{0.50\linewidth}
\centering
\subcaption{Training loss.}
\begin{minipage}[c][\dimexpr\ht\ablationlossbox+\dp\ablationlossbox\relax][c]{\linewidth}
\centering
\usebox{\ablationlossbox}
\end{minipage}
\end{subfigure}
\caption{Ablation on Qwen3-8B. (a) Mean accepted length by domain; (b) Training loss curve.}
\label{fig:ablation}
\vspace{-6pt}
\end{figure}

Figure~\ref{fig:ablation}(a) isolates each context component by removing it in turn, with up to 2,048 generated tokens per request. The four variants are trained for 5 epochs for a fair comparison. The recent KV window proves the most critical: without it, average accepted length falls from 4.69 to 3.71, confirming that token-level local detail is the primary driver of proposal quality. Removing the global summary or the boundary state causes smaller degradations, to 4.35 and 4.28 respectively, so both are secondary yet not redundant. The training dynamics in Figure~\ref{fig:ablation}(b) add a complementary view: removing the global summary produces pronounced transient loss spikes, whereas the full model descends smoothly. The summary thus contributes to training stability as well as proposal quality, since without it a bounded-context drafter has no reliable signal about the global prefix and must reconstruct it from local evidence alone.


Together, these results reveal a fundamental asymmetry between drafter and target. The target requires a full, precise state to verify; the drafter needs only enough orientation to propose. The recent window supplies immediate local grounding, while the global summary supplies a stable, low-resolution view of the entire prefix. Detailed long-context ablations are provided in Appendix~\ref{app:longcontext-ablation}. A sliding-window variant of DSpark also bounds the drafter's cost, but does so by discarding the distant prefix, and it trails \swift{} in both accepted length and throughput at every context length (Appendix~\ref{app:swa}).

\section{Related Work}\label{sec:related}

We first review the lossless verification framework that makes drafter quality a pure efficiency concern, then survey how autoregressive, recurrent, and block-diffusion drafters condition on the confirmed prefix.

\paragraph{Lossless Speculative Decoding}\label{sec:related-lossless}
\citet{stern2018blockwise} first showed that a drafter can propose multiple future tokens at once and that a Target model can verify them, accepting the longest prefix consistent with exact greedy decoding.
Speculative sampling extends this to stochastic generation: \citet{leviathan2023fast} use rejection and residual correction to preserve the Target distribution exactly.
Because the Target's verification preserves the output distribution regardless of proposal quality, the drafter determines only efficiency.
How it conditions on the confirmed prefix is therefore a design choice that determines how its cost scales with the prefix length.
\paragraph{Autoregressive Drafters}\label{sec:related-ar}
The standard construction uses a smaller autoregressive language model whose own KV cache grows with the prefix.
Later methods strengthen proposals by incorporating Target representations: EAGLE and EAGLE-3 reuse Target hidden features~\citep{li2024eagle,li2026eagle,hui2026peagle}, and ReDrafter conditions a recurrent proposer on the Target state~\citep{cheng2024redrafter}.
LongSpec bounds its own draft cache using windowed self-attention, but it still cross-attends to the Target's full KV history~\citep{yang2026longspec}.
Bounding the drafter's own cache does not remove this growing read, and these drafters still advance one position at a time.

\paragraph{Block-Parallel and Block-Diffusion Drafting}\label{sec:related-block}
Block-parallel methods take a different approach: instead of drafting tokens one by one, \citet{xiao2024parallelspec} and \citet{an2025pard} predict several future positions at once.
Medusa attaches independent prediction heads to the Target for fixed-horizon parallel proposals~\citep{cai2024medusa,wertheimer2024speculators}.
Each head conditions only on the Target's last hidden state, making its cost prefix-independent but limiting it to single-position predictions that do not attend to the prefix.
DFlash replaces these independent heads with a block-diffusion backbone that predicts all masked positions jointly and injects Target features from the entire confirmed prefix into every draft layer~\citep{chen2026dflash,li2025diffuspec}.
This yields more expressive proposals, but makes every draft layer process a context sequence that grows with the prefix.
Orthrus shares the Target KV cache to eliminate duplicate storage, but its diffusion view still attends to the complete cache, leaving prefix-length context computation intact~\citep{nguyen2026orthrus}.

\looseness=-1\relax Subsequent work improves along other dimensions while retaining the same growing context interface.
DFlare strengthens Target-to-drafter feature fusion~\citep{zhang2026dflare}; DSpark, Domino, JetSpec and xPress restore causal dependence within the proposed block~\citep{cheng2026dspark,hu2026jetspec,huang2026domino,wang2026xpress}; DDTree organizes the position-wise distributions into a draft tree~\citep{agrawal2025draftgraphs,gao2026progressivetree,ringel2026ddtree,wang2026presto}; and AdaFlash, like DSpark, schedules the proposal horizon with a dynamic confidence head~\citep{cheng2026dspark,qian2026adaflash}.
Each of these improves feature conditioning, within-block coherence, candidate coverage, or the proposal horizon, yet each keeps the context that proposals read tied to the prefix. The DSpark drafters co-deployed with DeepSeek-V4 adopt a 128-token sliding window~\citep{cheng2026dspark}, which bounds this context by truncating it to the most recent tokens; Appendix~\ref{app:swa} compares the two routes to a fixed cost.
\swift{} changes this remaining axis: it is, to our knowledge, the first block-diffusion drafter to replace the growing context with a fixed-size, multiscale view of the Target state that still covers the entire prefix.
Target verification is consequently the only full-prefix traversal in each round; all additional context processing, proposal computation, and drafter state remain fixed as the context grows.

\section{Conclusion and Future Directions}\label{sec:conclusion}

This work establishes that, during decoding, a speculative drafter's cost can be made entirely independent of the prefix length while maintaining competitive accepted lengths. Because verification guarantees output correctness, the drafter needs only enough context to produce useful proposals, and we show that this context can be extracted from the target's state at fixed cost.

The end-to-end results bear out the value of this fixed cost. \swift{} attains the highest throughput across model scales even though the strongest prefix-scaling drafter reaches slightly longer accepted lengths, showing that low proposal overhead outweighs marginal acceptance gains. This advantage grows with context length and serving concurrency because the drafter's cost remains constant while the baselines' costs grow. Ablations further show that a compact, lossy view of the prefix provides sufficient orientation for competitive proposals. Together, these results identify accepted tokens per unit of proposal overhead as the quantity drafting design should optimize.

Future directions include evaluating the fixed-cost interface with larger target models and longer context windows to assess its effectiveness as model scale and context demands increase. The interface developed here offers one viable realization of fixed-cost drafting; alternative interface designs and drafter architectures within this framework may further improve proposal quality while retaining prefix-independent cost. Beyond proposal design, exploring tree-structured and multi-drafter verification schemes could help increase the number of accepted tokens per round while preserving the $O(1)$ drafting cost. On the application side, reinforcement-learning rollouts offer a promising setting for these extensions, as many long trajectories run under tight latency budgets and a constant proposal overhead across trajectory length could compound throughput gains.

\section*{AI Use of Generative Models}\label{sec:ai-disclosure}

We used generative AI tools to assist with the research workflow: developing and debugging code, analysing experimental results, and preparing the manuscript, tables, and figures.
The authors reviewed and verified all AI-assisted work and take full responsibility for the final content of this work, including all text, claims, and artefacts.

\bibliography{iclr2027_conference}
\bibliographystyle{iclr2027_conference}

\clearpage
\appendix
\section*{\LARGE \centering Appendix}
\section{Method and Algorithmic Details}\label{app:method-details}

This section details the proposal computation and the incremental maintenance of the global summary.

\subsection{Drafter Forward Pass}\label{app:drafter-forward}

We first describe how the drafter turns the three fixed-size context views into one block proposal.

\begin{figure}[h]
\centering
\vspace{6pt}
\begin{tcolorbox}[colback=white,colframe=codeborder,boxrule=0.7pt,arc=4pt,
                  boxsep=0pt,left=9pt,right=9pt,top=9pt,bottom=9pt,
                  before skip=0pt,after skip=0pt,width=\linewidth]
\includegraphics[width=\linewidth]{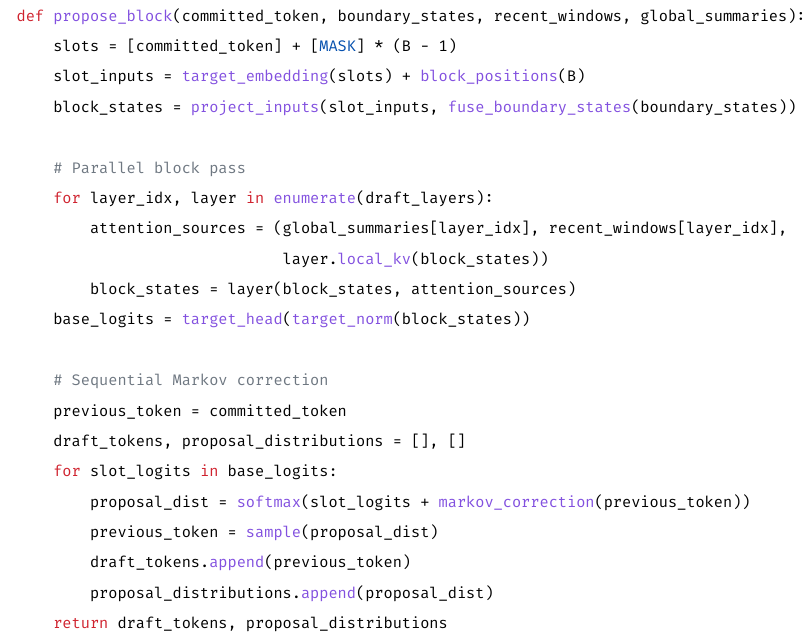}
\end{tcolorbox}
\vspace{5pt}
\caption{One block proposal by the \swift{} drafter.}
\label{fig:drafter-forward}
\vspace{6pt}
\end{figure}

Figure~\ref{fig:drafter-forward} summarizes one proposal step after the Target has cached $x_{1:t}$, with $\tok{a}=x_{t+1}$ as the next committed token. The drafter takes the selected-layer boundary states $\{\bm h_t^\ell\}_{\ell\in\gL}$, recent Target KV windows, and global summary pairs $(\bm G,\bm M_t)$, all ordered by selected target layer.

Input slot 0 supplies the base logits for $y_1$. All slots pass through the draft layers in parallel, with one attention normalization across the three sources and bidirectional attention among local slots. The Target embedding, final normalization, and vocabulary head remain frozen. Local block KV is discarded after drafting; only the Markov correction and sampling proceed sequentially. The returned distributions are used in verification, after which retained Target KV rows update the summary as described next.

\subsection{Incremental Global Context Summary}\label{app:summary-update}

We then derive the update that folds newly retained rows into the global summary at constant cost.

For a fixed summary query $\bm{g}$, let $s_j=\bm{g}^\top \bm{k}_j/\sqrt d$ be its attention score for Target key $\bm{k}_j$.
For any nonempty set of KV rows $A$, define its summary value and log-normalizer as
\begin{equation}
\eta_A=\log\sum_{j\in A}\exp(s_j),
\qquad
\bm{m}_A=\sum_{j\in A}\exp(s_j-\eta_A)\bm{v}_j.
\end{equation}
Let $A$ contain the previously processed prefix and $N$ the newly retained rows.
Their attention statistics merge as
\begin{equation}
\begin{aligned}
\eta &= \log\bigl(\exp(\eta_A)+\exp(\eta_N)\bigr),\\
\bm{m} &= \exp(\eta_A-\eta)\bm{m}_A+\exp(\eta_N-\eta)\bm{m}_N.
\end{aligned}
\label{eq:summary-update}
\end{equation}
The merged value $\bm{m}$ equals attention over $A\cup N$; an empty new set leaves the state unchanged.
The log-sum-exp operations are evaluated with max subtraction for numerical stability.
Applying this merge independently to all summary queries recovers \eqref{eq:global-summary} while storing only one value vector and one log-normalizer per query.
Since the queries and historical Target KV remain fixed, only the statistics over $N$ need to be computed in each round.

\section{Model Architecture and Training}\label{app:implementation}

This section records the architecture and training settings shared by all reported drafters.

\subsection{Drafter Architecture}\label{app:arch}

We specify the drafter depth, its context-view sizes, and the Target layers it reads.

\swift{} employs a lightweight decoder comprising 5 Qwen3-style blocks that operate in parallel to propose 7 tokens. To maintain architectural consistency, all drafters adhere to the standard Qwen3 hyperparameters for a given hidden dimension $d$ (e.g., head dimension and intermediate width). The architecture is designed as an efficient distilled version of the Target, integrating a global context summary, a recent KV window, and a small local workspace. We extract context views from Target layers $\{0,9,18,26,35\}$ for Qwen3-4B and 8B, and $\{0,10,20,29,39\}$ for Qwen3-14B. Across all scales, the global context summary contains $R=16$ entries per selected layer and attention head, and the recent KV window contains $W=256$ entries per selected layer. We reuse the Target's embedding and final LM head to ensure vocabulary consistency.

\subsection{Training Procedure}\label{app:training}

We list the data, objective, and optimization settings used to train each drafter.

We train the \swift{} drafters with the objective in \eqref{eq:training-objective} on the Open-PerfectBlend dataset~\citep{labonne2024perfectblend}, which contains approximately 1.42M sessions. For each target scale, we regenerate the assistant responses in a non-thinking mode using the frozen Target model ($\text{temperature}=0.7, \text{top-p}=0.8, \text{top-k}=20$). During training, we limit the total sequence length to 4096 tokens and randomly sample up to 512 anchors per sequence to optimize the distillation loss across multiple positions in each generated response.

\paragraph{Summary queries.}
Each selected layer and attention head has an independent bank of $R$ summary queries, initialized from $\mathcal{N}(\bm{0},d^{-1}\bm{I})$ and jointly optimized with the drafter under the same training objective, where $d$ is the head dimension. The queries are shared across examples and remain fixed during inference, enabling the incremental summary updates in Appendix~\ref{app:summary-update}.

The main drafters are trained for 10 epochs and the ablation variants for 5 epochs, all with a global batch size of 512. We use the AdamW optimizer with BF16 mixed-precision and FP32 master weights, a peak learning rate of $6 \times 10^{-4}$, and a cosine decay scheduler with a 4\% linear warmup. For all model scales, we employ a data-parallel (DP) degree of 8 with gradient accumulation steps of 64.

\section{Evaluation Protocol and Dataset Details}\label{app:evaluation-details}

This section documents the long-context workloads and the measurement protocol behind the reported numbers.

\subsection{Long-Context Evaluation Datasets}\label{app:codespan}

We describe the composition of the three long-context workloads.

\paragraph{LongSpec.}
We construct continuation tasks from the code, arXiv, and book subsets of LongSpec's released training corpus~\citep{yang2026longspec}, using 32,768-token inputs.
These subsets form the LongSpec 32K evaluation workload.
Our drafting-cost analysis uses a prefix from the code subset.

\paragraph{CodeSpan.}
CodeSpan comprises 32 source-code continuation examples drawn from 32 distinct files across 17 open-source projects, including LLVM, GCC, Linux, and PostgreSQL.
We select sufficiently long files, with at most four files per project.
The collection is predominantly C/C++ (28 files), with additional Python, TypeScript, and Yacc grammar files.
Each example uses a contiguous prefix of a single file, without concatenation or repetition.
Using the Qwen3-8B tokenizer, we choose the prefix length so that the complete input, including the continuation instruction and chat template, contains 65,536 tokens.

\paragraph{LongSWE-Bench.}
LongSWE-Bench~\citep{rando2025longcodebench} evaluates repository-level bug fixing with long code contexts.
We select inputs that fit within a 128K total context budget while reserving space for up to 8,192 generated tokens.
The same evaluation inputs and ordering are shared across methods and random seeds.

\subsection{Measurement Protocol}\label{app:measurement}

We define how throughput, accepted length, and time per output token are measured.

\paragraph{Throughput and accepted length.}
We measure steady-state serving performance over a 120-second interval after a 30-second warm-up, continuously admitting new requests as others finish.
Each method samples its own responses independently.
Throughput counts all output tokens emitted during this interval.
For each seed, $\tau$ is computed by dividing the number of emitted tokens after the first token by the number of verification request-rounds, including bonus or residual tokens and accounting for sequence termination.
We take arithmetic means over seeds, then compute speedup as the ratio of a method's mean throughput to the autoregressive baseline's mean throughput.
For the long-context component ablations in Table~\ref{tab:longcontext-ablation}, drafters are trained for five epochs. Entries report the mean and standard deviation over ten seeds, with variants evaluated on the same host within each seed.

\paragraph{Time per output token.}
TPOT is the total post-first-token request time overlapping the measurement interval divided by the number of subsequent output tokens emitted within that interval.
The numerator includes the observed time of requests still active at the end of measurement.
This definition excludes time to first token while retaining scheduling and concurrent prefill effects.

\section{Additional Decoding Results}\label{app:additional-results}

This section collects results that complement the main evaluation.

\subsection{Greedy-Decoding Results}\label{app:accepted-length}

We first check whether the efficiency advantage depends on sampling.

Table~\ref{tab:accepted-length-supp} presents greedy-decoding results ($T=0$) with target TP1, concurrency 32, and up to 2,048 generated tokens per request, complementing the sampling-based evaluation in Table~\ref{tab:main-results}. The performance trends here closely mirror those observed at $T=1$, with \swift{} consistently achieving the highest speedups across all model scales and benchmarks. This consistency demonstrates that the efficiency gains of \swift{} are a structural advantage independent of the decoding strategy, further validating that the $O(1)$ proposal cost is the primary driver of end-to-end performance regardless of the temperature setting.

\begin{table}[htbp]
\centering
\captionsetup{skip=6pt}
\caption{Greedy decoding across model scales ($T=0$). Entries report accepted length and throughput speedup using the conventions of Table~\ref{tab:main-results}.}
\label{tab:accepted-length-supp}
\scriptsize
\setlength{\tabcolsep}{0.75pt}
\renewcommand{\arraystretch}{1.28}
\begin{tabularx}{\linewidth}{@{}c>{\hspace{3pt}}l*{9}{>{\centering\arraybackslash}X}@{}}
\toprule
\multirow{2}{*}{Target} & \multirow{2}{*}{Drafter} & \multicolumn{3}{c}{\textbf{Math}} & \multicolumn{3}{c}{\textbf{Code}} & \multicolumn{2}{c}{\textbf{Chat}} & \textbf{Overall} \\
\cmidrule(lr){3-5} \cmidrule(lr){6-8} \cmidrule(lr){9-10} \cmidrule(lr){11-11}
 & & GSM8K & MATH-500 & AIME25 & MBPP & HumanEval & LCB & MT-Bench & Alpaca & Avg. \\
\midrule
\multirow{4}{*}{Q3-4B} & EAGLE-3 & 5.46 (1.55) & 5.23 (1.68) & 4.72 (1.68) & 4.12 (1.24) & 4.56 (1.40) & 4.92 (1.60) & 2.70 (0.89) & 2.48 (0.82) & 4.27 (1.36) \\
 & DFlash & 5.93 (2.21) & 5.87 (2.54) & 5.28 (2.57) & 4.88 (1.90) & 5.21 (2.12) & 5.34 (2.25) & 3.44 (1.51) & 3.30 (1.43) & 4.90 (2.07) \\
 & DSpark & 6.33 (2.26) & 6.30 (2.64) & 5.77 (2.73) & 5.38 (2.01) & 5.68 (2.24) & 5.71 (2.33) & 3.84 (1.62) & 3.67 (1.53) & 5.34 (2.17) \\
\rowcolor{tablehighlight}
\cellcolor{white} & \swift{} & 6.14 (\textbf{2.41}) & 6.07 (\textbf{2.85}) & 5.44 (\textbf{2.94}) & 5.27 (\textbf{2.17}) & 5.37 (\textbf{2.37}) & 5.44 (\textbf{2.47}) & 3.70 (\textbf{1.76}) & 3.60 (\textbf{1.68}) & 5.13 (\textbf{2.33}) \\
\midrule
\multirow{4}{*}{Q3-8B} & EAGLE-3 & 5.58 (1.68) & 5.34 (1.79) & 4.88 (1.78) & 4.30 (1.35) & 4.68 (1.51) & 5.11 (1.66) & 2.92 (1.00) & 2.75 (0.94) & 4.44 (1.46) \\
 & DFlash & 6.00 (2.37) & 5.91 (2.66) & 5.41 (2.69) & 4.89 (2.00) & 5.27 (2.25) & 5.34 (2.19) & 3.41 (1.56) & 3.32 (1.51) & 4.94 (2.15) \\
 & DSpark & 6.46 (2.50) & 6.38 (2.82) & 5.89 (2.88) & 5.47 (2.17) & 5.81 (2.41) & 5.77 (2.30) & 3.82 (1.70) & 3.73 (1.65) & 5.42 (2.30) \\
\rowcolor{tablehighlight}
\cellcolor{white} & \swift{} & 6.24 (\textbf{2.59}) & 6.14 (\textbf{2.96}) & 5.57 (\textbf{3.01}) & 5.32 (\textbf{2.28}) & 5.50 (\textbf{2.49}) & 5.53 (\textbf{2.37}) & 3.65 (\textbf{1.78}) & 3.60 (\textbf{1.74}) & 5.19 (\textbf{2.40}) \\
\midrule
\multirow{4}{*}{Q3-14B} & EAGLE-3 & 5.55 (1.90) & 5.23 (1.97) & 4.69 (1.89) & 4.09 (1.47) & 4.51 (1.63) & 4.90 (1.71) & 2.82 (1.08) & 2.68 (1.03) & 4.31 (1.59) \\
 & DFlash & 6.01 (2.59) & 5.90 (2.85) & 5.33 (2.77) & 4.88 (2.19) & 5.22 (2.39) & 5.21 (2.16) & 3.42 (1.67) & 3.32 (1.62) & 4.91 (2.28) \\
 & DSpark & 6.47 (2.70) & 6.37 (2.99) & 5.81 (2.94) & 5.47 (2.36) & 5.77 (2.54) & 5.66 (2.26) & 3.89 (1.83) & 3.76 (1.77) & 5.40 (2.43) \\
\rowcolor{tablehighlight}
\cellcolor{white} & \swift{} & 6.32 (\textbf{2.74}) & 6.21 (\textbf{3.10}) & 5.54 (\textbf{3.00}) & 5.39 (\textbf{2.46}) & 5.52 (\textbf{2.57}) & 5.46 (\textbf{2.29}) & 3.79 (\textbf{1.89}) & 3.68 (\textbf{1.83}) & 5.24 (\textbf{2.49}) \\
\bottomrule
\end{tabularx}
\end{table}

\subsection{Long-Context Performance Across Model Scales}\label{app:longcontext-scales}

We then check whether the long-context advantage is specific to the 8B target.

Table~\ref{tab:longcontext-scales} extends the Qwen3-8B evaluation to 4B and 14B. Across all three model scales, \swift{} achieves the highest average TPS and lowest average TPOT on each benchmark.

\begin{table}[htbp]
\centering
\captionsetup{skip=6pt}
\caption{Long-context decoding on Qwen3-4B and 14B, using the evaluation protocol and reporting conventions of Table~\ref{tab:long-context}.}
\label{tab:longcontext-scales}
\small
\renewcommand{\arraystretch}{1.12}
\setlength{\tabcolsep}{3pt}
\begin{tabularx}{\linewidth}{@{}c>{\hspace{3pt}}l*{6}{>{\centering\arraybackslash}X}@{}}
\toprule
\multirow{2}{*}{Target} & \multirow{2}{*}{Method} & \multicolumn{2}{c}{LongSpec 32K} & \multicolumn{2}{c}{CodeSpan 64K} & \multicolumn{2}{c}{LSWE 128K} \\
\cmidrule(lr){3-4} \cmidrule(lr){5-6} \cmidrule(lr){7-8}
 & & TPS $\uparrow$ & TPOT $\downarrow$ & TPS $\uparrow$ & TPOT $\downarrow$ & TPS $\uparrow$ & TPOT $\downarrow$ \\
\midrule
\multirow{5}{*}{Q3-4B} & Vanilla & 1250.1 & 12.7 & 864.4 & 18.2 & 185.1 & 83.0 \\
 & EAGLE-3 & 1810.5 & 8.7 & 1197.2 & 13.1 & 202.0 & 77.4 \\
 & DFlash & 1453.1 & 10.9 & 748.9 & 21.1 & 175.6 & 88.7 \\
 & DSpark & 1761.8 & 8.9 & 884.0 & 17.9 & 184.0 & 83.8 \\
\rowcolor{tablehighlight}
\cellcolor{white} & \swift{} & \textbf{2274.6} & \textbf{6.9} & \textbf{1497.0} & \textbf{10.4} & \textbf{233.2} & \textbf{66.2} \\
\midrule
\multirow{5}{*}{Q3-14B} & Vanilla & 951.8 & 16.6 & 615.7 & 25.5 & 115.1 & 133.6 \\
 & EAGLE-3 & 964.4 & 16.4 & 628.1 & 25.0 & 112.0 & 137.0 \\
 & DFlash & 1248.8 & 12.5 & 832.9 & 18.8 & 112.3 & 136.4 \\
 & DSpark & 1416.2 & 11.0 & 945.1 & 16.5 & 121.9 & 127.8 \\
\rowcolor{tablehighlight}
\cellcolor{white} & \swift{} & \textbf{1464.2} & \textbf{10.7} & \textbf{1002.8} & \textbf{15.5} & \textbf{133.5} & \textbf{115.3} \\
\bottomrule
\end{tabularx}
\end{table}

\Needspace{28\baselineskip}

\subsection{Long-Context Component Ablations}\label{app:longcontext-ablation}

We evaluate the contribution of each context component on workloads spanning 32K--128K tokens. Table~\ref{tab:longcontext-ablation} reports accepted length and serving performance for the full drafter and variants that remove one component at a time.

\begin{table}[htbp]
\centering

\caption{Long-context component ablations. All other settings follow Table~\ref{tab:long-context}.}
\label{tab:longcontext-ablation}
\scriptsize
\renewcommand{\arraystretch}{1.18}
\setlength{\tabcolsep}{2pt}

\begin{tabularx}{\linewidth}{@{}l*{6}{>{\centering\arraybackslash}X}@{}}
\multicolumn{7}{c}{(a) Accepted length $\tau\,\uparrow$} \\
\toprule
Variant & \multicolumn{2}{c}{LongSpec 32K} & \multicolumn{2}{c}{CodeSpan 64K} & \multicolumn{2}{c}{LSWE 128K} \\
\midrule
\rowcolor{tablehighlight}
\swift{} & \multicolumn{2}{c}{\textbf{2.88} $\pm$ 0.11} & \multicolumn{2}{c}{\textbf{3.40} $\pm$ 0.11} & \multicolumn{2}{c}{\textbf{2.53} $\pm$ 0.22} \\
w/o global summary & \multicolumn{2}{c}{2.70 $\pm$ 0.09} & \multicolumn{2}{c}{3.18 $\pm$ 0.11} & \multicolumn{2}{c}{2.25 $\pm$ 0.06} \\
w/o boundary state & \multicolumn{2}{c}{2.44 $\pm$ 0.06} & \multicolumn{2}{c}{2.82 $\pm$ 0.10} & \multicolumn{2}{c}{2.08 $\pm$ 0.10} \\
w/o recent KV window & \multicolumn{2}{c}{1.90 $\pm$ 0.07} & \multicolumn{2}{c}{1.77 $\pm$ 0.04} & \multicolumn{2}{c}{1.76 $\pm$ 0.08} \\
\bottomrule
\addlinespace[6pt]
\multicolumn{7}{c}{(b) Serving performance} \\
\toprule
\multirow{2}{*}{Variant} & \multicolumn{2}{c}{LongSpec 32K} & \multicolumn{2}{c}{CodeSpan 64K} & \multicolumn{2}{c}{LSWE 128K} \\
\cmidrule(lr){2-3} \cmidrule(lr){4-5} \cmidrule(lr){6-7}
 & TPS $\uparrow$ & TPOT $\downarrow$ & TPS $\uparrow$ & TPOT $\downarrow$ & TPS $\uparrow$ & TPOT $\downarrow$ \\
\midrule
\rowcolor{tablehighlight}
\swift{} & \textbf{1597.3} $\pm$ 77.0 & \textbf{9.79} $\pm$ 0.46 & \textbf{1527.2} $\pm$ 103.8 & \textbf{10.26} $\pm$ 0.67 & \textbf{208.4} $\pm$ 45.3 & \textbf{75.39} $\pm$ 13.98 \\
w/o global summary & 1514.4 $\pm$ 77.8 & 10.36 $\pm$ 0.52 & 1462.3 $\pm$ 90.5 & 10.74 $\pm$ 0.65 & 183.5 $\pm$ 10.6 & 82.90 $\pm$ 4.89 \\
w/o boundary state & 1460.5 $\pm$ 76.9 & 10.75 $\pm$ 0.55 & 1391.6 $\pm$ 73.8 & 11.29 $\pm$ 0.58 & 177.4 $\pm$ 19.2 & 86.44 $\pm$ 8.85 \\
w/o recent KV window & 1146.3 $\pm$ 58.9 & 13.74 $\pm$ 0.71 & 976.3 $\pm$ 48.5 & 16.20 $\pm$ 0.79 & 178.1 $\pm$ 23.9 & 86.88 $\pm$ 11.96 \\
\bottomrule
\end{tabularx}
\end{table}

\subsection{Comparison with Sliding-Window Drafting}\label{app:swa}

We compare fixed-cost drafting with the most direct way to bound a drafter's cost: a sliding window, as adopted by the DSpark drafters co-deployed with DeepSeek-V4~\citep{cheng2026dspark}. We construct its Qwen3-8B counterpart, DSpark-SWA, by restricting the released DSpark checkpoint to a 256-token window at inference, covering as many recent tokens as the recent window of \swift{} but nothing beyond them.

\begin{table}[htbp]
\centering
\captionsetup{skip=6pt}
\caption{Comparison with sliding-window drafting on Qwen3-8B. We report accepted length $\tau$ and TPS (tokens/s); Main Benchmarks averages the eight benchmarks of Table~\ref{tab:main-results}. Bold marks the best.}
\label{tab:swa}
\small
\renewcommand{\arraystretch}{1.18}
\setlength{\tabcolsep}{3pt}
\begin{tabularx}{\linewidth}{l*{8}{>{\centering\arraybackslash}X}}
\toprule
\multirow{2}{*}{Method} & \multicolumn{2}{c}{Main Benchmarks} & \multicolumn{2}{c}{LongSpec 32K} & \multicolumn{2}{c}{CodeSpan 64K} & \multicolumn{2}{c}{LSWE 128K} \\
\cmidrule(lr){2-3} \cmidrule(lr){4-5} \cmidrule(lr){6-7} \cmidrule(lr){8-9}
 & $\tau$ & TPS & $\tau$ & TPS & $\tau$ & TPS & $\tau$ & TPS \\
\midrule
DSpark & \textbf{5.00} & 6741.4 & \textbf{3.36} & 1736.6 & 2.81 & 1217.3 & 1.87 & 190.9 \\
DSpark-SWA & 4.72 & 6474.9 & 2.93 & 1714.6 & 2.80 & 1373.6 & 2.33 & 196.3 \\
\rowcolor{tablehighlight}
\swift{} & 4.78 & \textbf{6987.2} & 3.23 & \textbf{1917.6} & \textbf{3.47} & \textbf{1571.8} & \textbf{2.49} & \textbf{212.7} \\
\bottomrule
\end{tabularx}
\end{table}

Truncating the prefix leaves DSpark-SWA behind \swift{} in both accepted length and throughput on every workload (Table~\ref{tab:swa}). \swift{} instead draws its context from the Target's own state, keeping the entire prefix in view at a fixed cost.

\subsection{Performance Across Concurrency Levels}\label{app:concurrency}

We next check how the advantage changes with serving load.

Table~\ref{tab:concurrency-supp} extends the Qwen3-8B evaluation to concurrency levels 8, 32, and 128, using target TP1, $T=1$, and up to 2,048 generated tokens per request. The concurrency-32 results reproduce the corresponding entries in Table~\ref{tab:main-results}. We observe a clear trend: while DSpark is competitive at low concurrency (e.g., $C=8$), \swift{} becomes increasingly dominant as the number of concurrent requests grows. At $C=128$, \swift{} maintains a significant throughput lead over all baselines. This scaling behavior stems from the minimal state overhead of \swift{}. Unlike existing drafters that must manage and access large KV caches for each concurrent request, \swift{}'s fixed-cost drafting mechanism avoids the memory-bandwidth bottleneck associated with scaling the drafter's state. Consequently, \swift{} is exceptionally well-suited for high-concurrency serving environments where maximizing total throughput is critical.

\begin{table}[htbp]
\centering
\captionsetup{skip=6pt}
\caption{Concurrency scaling on Qwen3-8B ($C$: concurrency). Entries follow Table~\ref{tab:main-results}, with speedups relative to autoregressive decoding at the same concurrency.}
\label{tab:concurrency-supp}
\scriptsize
\setlength{\tabcolsep}{0.75pt}
\renewcommand{\arraystretch}{1.28}
\begin{tabularx}{\linewidth}{@{}c>{\hspace{3pt}}l*{9}{>{\centering\arraybackslash}X}@{}}
\toprule
\multirow{2}{*}{$C$} & \multirow{2}{*}{Drafter} & \multicolumn{3}{c}{\textbf{Math}} & \multicolumn{3}{c}{\textbf{Code}} & \multicolumn{2}{c}{\textbf{Chat}} & \textbf{Overall} \\
\cmidrule(lr){3-5} \cmidrule(lr){6-8} \cmidrule(lr){9-10} \cmidrule(lr){11-11}
 & & GSM8K & MATH-500 & AIME25 & MBPP & HumanEval & LCB & MT-Bench & Alpaca & Avg. \\
\midrule
\multirow{4}{*}{8} & EAGLE-3 & 5.23 (1.92) & 4.70 (1.82) & 4.01 (1.61) & 3.91 (1.48) & 4.34 (1.65) & 4.18 (1.58) & 2.72 (1.06) & 2.54 (0.99) & 3.95 (1.51) \\
 & DFlash & 5.38 (2.68) & 4.92 (2.64) & 4.07 (2.27) & 4.37 (2.25) & 4.68 (2.44) & 4.50 (2.27) & 3.09 (1.66) & 2.99 (1.60) & 4.25 (2.23) \\
 & DSpark & 6.16 (\textbf{2.98}) & 5.78 (\textbf{3.03}) & 5.00 (\textbf{2.72}) & 5.19 (\textbf{2.59}) & 5.47 (\textbf{2.78}) & 5.12 (\textbf{2.49}) & 3.67 (\textbf{1.92}) & 3.60 (\textbf{1.87}) & 5.00 (\textbf{2.55}) \\
\rowcolor{tablehighlight}
\cellcolor{white} & \swift{} & 5.96 (2.96) & 5.57 (2.98) & 4.69 (2.63) & 5.03 (2.54) & 5.14 (2.69) & 4.86 (2.44) & 3.51 (1.89) & 3.46 (1.82) & 4.78 (2.49) \\
\midrule
\multirow{4}{*}{32} & EAGLE-3 & 5.26 (1.49) & 4.75 (1.50) & 4.02 (1.37) & 3.92 (1.16) & 4.32 (1.30) & 4.17 (1.28) & 2.67 (0.85) & 2.54 (0.82) & 3.96 (1.22) \\
 & DFlash & 5.36 (1.96) & 4.91 (2.02) & 4.07 (1.83) & 4.36 (1.66) & 4.67 (1.82) & 4.44 (1.68) & 3.09 (1.28) & 3.00 (1.24) & 4.24 (1.69) \\
 & DSpark & 6.17 (2.17) & 5.80 (2.32) & 4.99 (2.19) & 5.18 (1.89) & 5.48 (2.06) & 5.09 (1.84) & 3.69 (1.47) & 3.59 (1.42) & 5.00 (1.92) \\
\rowcolor{tablehighlight}
\cellcolor{white} & \swift{} & 5.93 (\textbf{2.24}) & 5.57 (\textbf{2.42}) & 4.72 (\textbf{2.27}) & 5.03 (\textbf{1.98}) & 5.16 (\textbf{2.11}) & 4.83 (\textbf{1.88}) & 3.53 (\textbf{1.54}) & 3.46 (\textbf{1.49}) & 4.78 (\textbf{1.99}) \\
\midrule
\multirow{4}{*}{128} & EAGLE-3 & 5.24 (1.10) & 4.73 (1.06) & 4.02 (1.02) & 3.91 (0.82) & 4.33 (0.92) & 4.14 (0.99) & 2.69 (0.60) & 2.54 (0.58) & 3.95 (0.88) \\
 & DFlash & 5.37 (1.33) & 4.92 (1.30) & 4.09 (1.22) & 4.37 (1.07) & 4.67 (1.16) & 4.46 (1.19) & 3.09 (0.80) & 3.00 (0.79) & 4.25 (1.11) \\
 & DSpark & 6.17 (1.49) & 5.80 (1.50) & 4.98 (1.45) & 5.19 (1.25) & 5.48 (1.32) & 5.11 (1.29) & 3.68 (0.93) & 3.59 (0.93) & 5.00 (1.27) \\
\rowcolor{tablehighlight}
\cellcolor{white} & \swift{} & 5.95 (\textbf{1.63}) & 5.57 (\textbf{1.62}) & 4.72 (\textbf{1.58}) & 5.03 (\textbf{1.35}) & 5.16 (\textbf{1.41}) & 4.84 (\textbf{1.36}) & 3.52 (\textbf{1.01}) & 3.46 (\textbf{1.00}) & 4.78 (\textbf{1.37}) \\
\bottomrule
\end{tabularx}
\end{table}

\Needspace{14\baselineskip}
\subsection{Fixed-Request End-to-End Performance}\label{app:fixed-requests}

We finally report end-to-end completion time for a fixed set of requests.

\begin{table}[h]
\vspace{-0.5\baselineskip}
\centering
\captionsetup{skip=6pt}
\caption{Mean end-to-end completion time for 32 requests at concurrency 16, averaged over three seeds (seconds; lower is better).}
\label{tab:fixed-requests}
\footnotesize
\renewcommand{\arraystretch}{1.28}
\setlength{\tabcolsep}{3pt}
\begin{tabularx}{\linewidth}{l*{3}{>{\centering\arraybackslash}X}}
\toprule
Method & \shortstack{LongSpec 32K} & \shortstack{CodeSpan 64K} & \shortstack{LSWE 128K} \\
\midrule
EAGLE-3 & 82.61 & 175.67 & 127.39 \\
DFlash & 71.11 & 202.01 & 138.75 \\
DSpark & 63.66 & 153.60 & 116.23 \\
\rowcolor{tablehighlight}
\swift{} & \textbf{56.14} & \textbf{106.77} & \textbf{99.71} \\
\bottomrule
\end{tabularx}
\end{table}

\vspace*{.1in}

We complement the steady-state throughput measurements with the time required to complete a fixed set of 32 requests on each long-context benchmark.
All requests arrive together and are served with concurrency 16 using Qwen3-8B, Target TP4, $T=1$, and a maximum of 8,192 generated tokens.
Elapsed time includes prefill, queuing, decoding, and completion of the final requests, excluding model loading and warm-up.
Table~\ref{tab:fixed-requests} reports mean completion times.
Responses are sampled independently, so completion times reflect both serving efficiency and variation in generated response length.
\swift{} achieves the lowest mean completion time on all three workloads.

\end{document}